\documentclass[a4paper,fleqn]{cas-dc}

\usepackage{soul} 
\usepackage{color, xcolor}
\definecolor{light-gray}{gray}{0.90}

\usepackage[numbers]{natbib}

\usepackage{subfigure}
\usepackage{tabularx}
\usepackage{tikz}
\usepackage{pgfplots}
\usepackage[edges]{forest}
\usepackage{adjustbox}
\hypersetup{
    colorlinks=true,
    linkcolor=blue,
    urlcolor=blue,
    anchorcolor=blue,
    citecolor=blue
}
\usepackage{makecell}

\usepackage{tabularx}

\begin{document}

\shorttitle{Enhancing Temporal Awareness in LLMs for Temporal Point Processes}
\shortauthors{L. Chen \textit{et al.}}

\author[1]{Lili Chen}
\ead{lilichien3@gmail.com}
\address[1]{School of Computer Science and Technology, Tongji University, Shanghai 201804, China}

\author[2]{Wensheng Gan}
\ead{wsgan001@gmail.com}
\address[2]{College of Cyber Security, Jinan University, Guangzhou 510632, China}

\author[1]{Shuang Liang}
\cortext[cor1]{Corresponding author}
\ead{shuangliang@tongji.edu.cn}
\cormark[1]

\author[3]{Philip S. Yu}
\ead{psyu@uic.edu}
\address[3]{Department of Computer Science,  University of Illinois Chicago, Chicago, USA}

\title[mode = title]{Enhancing Temporal Awareness in LLMs for Temporal Point Processes}

\begin{abstract}
  Temporal point processes (TPPs) are crucial for analyzing events over time and are widely used in fields such as finance, healthcare, and social systems. These processes are particularly valuable for understanding how events unfold over time, accounting for their irregularity and dependencies. Despite the success of large language models (LLMs) in sequence modeling, applying them to temporal point processes remains challenging. A key issue is that current methods struggle to effectively capture the complex interaction between temporal information and semantic context, which is vital for accurate event modeling. In this context, we introduce TPP-TAL (Temporal Point Processes with Enhanced Temporal Awareness in LLMs), a novel plug-and-play framework designed to enhance temporal reasoning within LLMs. Rather than using the conventional method of simply concatenating event time and type embeddings, TPP-TAL explicitly aligns temporal dynamics with contextual semantics before feeding this information into the LLM. This alignment allows the model to better perceive temporal dependencies and long-range interactions between events and their surrounding contexts. Through comprehensive experiments on several benchmark datasets, it is shown that TPP-TAL delivers substantial improvements in temporal likelihood estimation and event prediction accuracy, highlighting the importance of enhancing temporal awareness in LLMs for continuous-time event modeling. The code is made available at \href{https://github.com/chenlilil/TPP-TAL} {https://github.com/chenlilil/TPP-TAL}.
\end{abstract}

\begin{keywords}
  temporal point process\\
  LLMs\\
  temporal awareness\\
  event forecasting\\
  time–semantic alignment
\end{keywords}

\maketitle

\section{Introduction}

In the real world, events such as user behavior \cite{cai2018modeling,gan2018extracting}, medical incidents \cite{cao2025interpretable}, social interactions \cite{khodadadi2018continuous}, and financial transactions \cite{mollaev2025multimodal} often occur at irregular time intervals \cite{liang2025targeted}. To effectively model the dynamic mechanisms behind such event sequences, Temporal Point Processes (TPPs) \cite{daley2003introduction, yang2024neuro} have long been a fundamental tool in event modeling. Traditional statistical TPP models (e.g., Hawkes processes) \cite{laub2015hawkes} typically assume a specific analytical form for the intensity function, which limits their ability to capture complex and high-dimensional event semantics. To overcome this limitation, a series of neural TPP models have emerged in recent years, which are capable of automatically learning temporal dependencies and the historical context between events. For example, RMTPP \cite{du2016recurrent} was the first to introduce recurrent neural networks (RNNs) to parameterize the intensity function, allowing the capture of long-term dependencies between events. NHP \cite{mei2017neural} builds on the Hawkes model by incorporating continuous-time latent state updates, improving its ability to adapt to irregular time intervals. THP \cite{zuo2020transformer} applies self-attention mechanisms to model event sequences, enhancing parallelization and global dependency modeling. SAHP \cite{zhang2020self} further integrates time-shifted positional encoding to capture the relative time differences between events. While these neural TPP models significantly enhance the flexibility and predictive performance of temporal modeling, they typically treat event types as discrete labels, overlooking the rich semantic information embedded in event texts. This limitation restricts their applicability in semantically rich scenarios.

In recent years, Large Language Models (LLMs) \cite{naveed2025comprehensive, gan2023model, wu2023multimodal} have made significant breakthroughs in semantic understanding and knowledge representation. As a result, the combination of LLMs with TPPs has gradually become a new research trend \cite{shi2023language, liu2025tpp, kong2025language}. These methods aim to jointly model temporal dependencies and semantic contexts within a unified framework, thereby improving event prediction and generation. For example, Time-LLM \cite{jin2023time} redefines time series forecasting as a language modeling problem by mapping time information into token sequences that LLMs can process, employing a "reprogramming" strategy. LAMP \cite{shi2023language} enables LLMs to make accurate event predictions from incomplete or ambiguous information using few-shot abductive reasoning, even with limited labeled data. TPP-LLM \cite{liu2025tpp} introduces temporal embeddings into a frozen LLM backbone through parameter-efficient fine-tuning (LoRA), allowing the model to capture both the temporal features of events and their semantic representations. Furthermore, Language-TPP \cite{kong2025language} introduces a unified framework that combines language modeling with TPPs. It encodes time differences or timestamps as time tokens, which are fed into LLMs along with event types and descriptions to predict event time, type, and intensity, eliminating the need for explicit intensity function modeling.

Despite significant advances in semantic and temporal modeling, current approaches still face major challenges in effectively integrating temporal and semantic information. First, most LLM-based TPP methods treat time features as passive inputs, either concatenating or adding them to semantic representations without employing effective cross-modal interaction mechanisms. This limits the ability of temporal information to actively influence semantic representations, resulting in insufficient exploration of the relationship between time and semantics. Second, existing frameworks often represent event time differences using sequence positions or time embeddings, but fail to incorporate fine-grained temporal modulation within the attention mechanism. As a result, the model struggles to learn intricate temporal relationships, especially long-range interactions and cross-event temporal patterns. Furthermore, the absence of sophisticated time-aware semantic learning mechanisms constrains the model’s ability to identify subtle shifts in event dynamics over time. Overall, while LLM-based TPP models have made progress in capturing semantic information, they have not fully harnessed the bidirectional interaction between temporal signals and semantic representations, thereby limiting their ability to model complex temporal data effectively.

To address these challenges, this paper proposes TPP-TAL (Temporal Point Processes with Enhanced Temporal Awareness in LLMs), a novel framework designed to enhance temporal modeling in LLM-based TPP models. Specifically, the framework introduces two plug-and-play modules that optimize both event-level and cross-event temporal modeling. Temporal Cross-Fusion (TCF), for example, employs a lightweight cross-attention mechanism to explicitly modulate the interaction between temporal and semantic information within individual events. This enables more precise temporal-semantic fusion, thereby enhancing event-level modeling. Additionally, the Multi-Scale Temporal Bias Transformer (MTBT) introduces per-head additive biases based on event-level time differences, increasing the model’s sensitivity to temporal variations and improving its ability to capture temporal dependencies across events by incorporating multi-scale temporal information. To conclude, the key contributions of this work are outlined below:

\begin{itemize}
    \item We introduce the TPP-TAL framework, which significantly enhances both event-level and cross-event temporal modeling, ensuring that temporal information effectively influences semantic representations at both the local and global scales.
    
    \item We propose the TCF and MTBT modules to improve the integration of temporal and semantic information in LLM-based TPP models. TCF uses a cross-attention mechanism to fuse temporal and semantic features within individual events, enhancing event-level modeling. MTBT introduces per-head temporal biases based on time differences, increasing sensitivity to temporal variations and improving the capture of temporal dependencies across events.

    \item Comprehensive experiments across various real-world datasets show that TPP-TAL surpasses current methods in event type and time prediction.
\end{itemize}

The remainder of this paper is organized as follows: In Section \ref{Related work}, we provide a detailed review of related work. In Section \ref{Methodology}, we describe the TPP-TAL framework and its components. Section \ref{Experiment} presents the experimental setup, results, and comparison of TPP-TAL with existing methods. Finally, Section \ref{conclusion} summarizes the paper and highlights potential directions for further study.

\section{Related work}   
\label{Related work}

\subsection{Neural Temporal Point Processes}

Classical TPPs (e.g., Poisson and Hawkes processes) rely on predefined intensity functions and strong assumptions about baselines and triggering mechanisms, which significantly limit their ability to model nonlinear effects, heterogeneous dynamics, and high-dimensional semantics. These models often assume simplistic, predefined structures for event occurrence and interactions, making them ill-suited for capturing the complexities inherent in real-world event sequences, especially those in dynamic, unpredictable, or evolving environments. In particular, traditional models fail to adapt to the intricacies of temporal dependencies and cross-event correlations that can vary over time, and they struggle to incorporate the rich semantic context that often plays a pivotal role in real-world data. 
    
In contrast, neural approaches represent a significant shift, replacing rigid, hand-crafted structures with learnable, flexible representations. For instance, RMTPP \cite{du2016recurrent} introduces RNN-based architectures to capture long-range temporal dependencies, allowing the model to account for varying inter-event dynamics. NHP \cite{mei2017neural} goes a step further by incorporating continuous-time latent states, providing a more adaptable framework to handle irregular event intervals, a critical feature for datasets with varying time gaps between events. Building upon these developments, a dual-LSTM framework \cite{xiao2017modeling} was proposed to jointly model both the TPP dynamics and external time-series covariates, resulting in substantial improvements in predictive accuracy, particularly for event sequences influenced by additional time-varying features. Moreover, intensity-free and distribution-learning variants \cite{omi2019fully, upadhyay2018deep} address the limitations of likelihood-based assumptions, offering greater flexibility in event generation by relaxing these rigid constraints. This enables models to better adapt to diverse types of event sequences, even when the underlying intensity function cannot be easily specified. Transformer-based architectures have further propelled neural TPPs, enhancing their ability to model complex temporal relationships. Models such as THP \cite{zuo2020transformer} and SAHP \cite{zhang2020self} leverage global self-attention mechanisms, allowing the model to capture long-range dependencies more effectively. Additionally, innovations in continuous-time attention mechanisms and efficient sampling \cite{gao2024rothp, li2023smurf} have improved both the parallelization and temporal sensitivity of these models, making them more scalable and suitable for large datasets. Recent advancements have introduced more sophisticated mechanisms, such as nonlinear attention mechanisms based on Fourier kernels \cite{zhu2021deep}, which allow the model to capture more complex, periodic patterns in event sequences. Sparse attention mechanisms \cite{li2023sparse}, on the other hand, aim to improve computational efficiency by reducing the attention complexity, thus making these models more scalable without sacrificing performance.

While neural TPPs have significantly advanced in terms of capturing temporal dependencies and improving predictive performance, many existing methods still struggle with fully incorporating contextual semantics. This gap highlights a crucial opportunity for integrating large language models, which are adept at learning and leveraging semantic representations, into the TPP framework. By jointly learning temporal and semantic representations, LLMs can capture richer event context and interactions, leading to more accurate event predictions across diverse domains, ranging from social media analytics to financial prediction. This integration promises to enhance the flexibility and robustness of TPP models, opening up new possibilities for real-world event prediction tasks.

\subsection{LLM-based TPPs}

LLMs have demonstrated strong capabilities in semantic understanding and knowledge representation, and their potential for temporal reasoning and event prediction has recently gained increasing attention, particularly in fields requiring dynamic and real-time decision-making. As a result, LLM-based temporal point processes (TPPs) have emerged as a promising paradigm for continuous-time event modeling \cite{liu2025tpp, kong2025language, xue2023prompt}. These approaches aim to jointly encode temporal and semantic information, enabling LLMs to capture both the semantic context and temporal dynamics of events within a unified representation space, which is crucial for modeling complex, time-sensitive systems. Among these efforts, Time-LLM \cite{jin2023time} formulates time-series forecasting as a language modeling problem by transforming temporal signals into token sequences and introducing temporal prompts, allowing LLMs to learn time dependencies in an autoregressive manner. LAMP \cite{shi2023language} extends this by leveraging the abductive reasoning capability of LLMs to infer event progressions from incomplete or uncertain observations through natural-language reasoning chains. TPP-LLM \cite{liu2025tpp} integrates temporal and semantic modeling through parameter-efficient fine-tuning, injecting temporal embeddings into a frozen LLM and introducing an auxiliary head for joint prediction of event time and type, improving both interpretability and prediction accuracy. Language-TPP \cite{kong2025language} builds upon these advances by encoding timestamps or time differences as time tokens and combining them with event descriptions, enabling end-to-end prediction of event time, type, and intensity within a unified framework. In addition, PromptTPP \cite{xue2023prompt} focuses on streaming event sequences, adopting a prompt-based learning strategy that dynamically constructs contextual prompts to represent temporal-semantic relations, enabling LLMs to perform incremental and real-time event prediction. Recently, Zhou et al. \cite{zhou2025advances} provided a comprehensive review of recent developments in LLM-based TPPs, highlighting that the integration of LLMs is transforming traditional probabilistic modeling into a unified time-semantic representation paradigm, opening new opportunities for interpretable and general continuous-time event modeling.
    
Overall, while LLM-based TPPs show great potential for event modeling, they still face two key challenges: temporal information is often treated passively through simple fusion with semantics, and current models rarely capture detailed temporal-semantic interactions. Achieving deeper integration of time and semantics within LLMs remains a central direction for future research, as this will unlock even greater potential for accurate and adaptive event prediction in real-world applications.

\subsection{Time-Aware Representation Learning}

Time-aware representation learning has become a critical research direction in temporal modeling and event understanding. Time not only reflects the chronological order of events but also characterizes the dynamic evolution of complex systems, influencing both short-term and long-term predictions. Developing models that can perceive temporal variations, capture event dependencies, and adapt to changing dynamics remains a central challenge in continuous-time modeling.

Early approaches used positional or sinusoidal encodings \cite{vaswani2017attention, zhang2020self} to represent temporal order. While these methods capture the relative positions between events, they struggle to model irregular time intervals and continuous temporal dynamics, which are common in real-world event sequences. To address these limitations, later studies introduced continuous-time modeling mechanisms, such as temporal bias functions and temporal decay kernels \cite{mei2017neural, zuo2020transformer, che2018recurrent}. These mechanisms adjust attention weights based on temporal distances between events, enhancing the model's capacity to recognize long-range dependencies and temporal regularities, making it more suitable for dynamic event prediction. In parallel, time-aware architectures have been extensively explored in sequence modeling and language understanding. Time-aware Transformers \cite{xu2021temporal, dhingra2022time} integrate explicit temporal differences into attention computation, enabling models to distinguish between recent and distant events, which is critical for time-sensitive decision-making tasks. Similarly, time-sensitive embeddings \cite{hou2024dycars, kazemi2019time2vec} dynamically adapt representations to model the gradual decay of event influence over time, providing a more nuanced view of event progression. Temporal gating mechanisms and dynamic memory structures \cite{mei2017neural} have also been proposed to enhance the modeling of long-term dependencies and improve the interpretability of temporal reasoning. Beyond unimodal learning, recent research has explored integrating temporal, spatial, and semantic information within unified embedding spaces to support cross-modal event understanding and prediction \cite{huang2025enhancing}. This integration allows for a more holistic approach to event modeling, where both temporal and contextual information can be leveraged simultaneously. However, temporal information is often treated as an auxiliary feature rather than an intrinsic component of semantic representation, limiting the development of genuinely time-aware embeddings.
    
Existing research consistently highlights that explicit alignment between temporal dynamics and contextual semantics is essential for advancing temporal reasoning. When temporal and semantic information are jointly modeled within a shared representation space, models can more effectively capture event evolution, underlying causal dependencies, and the intricate interactions between time and context. Motivated by this insight, this paper proposes a unified architecture to enhance temporal awareness in large language models, improving fine-grained temporal reasoning and achieving better performance in continuous-time event modeling. This integration of time-aware semantic representations holds the potential to unlock more accurate and interpretable models for event prediction tasks.

\section{Methodology} \label{Methodology}
This section introduces the proposed TPP-TAL framework. We first define the basic concepts of continuous-time event modeling, followed by a description of the overall architecture and the two key modules that integrate temporal and semantic reasoning within LLMs. Finally, we explain how these modules are incorporated into the LLM input and outline the joint training objective used to optimize both temporal and semantic components.

\subsection{Problem Formulation}

We consider a continuous-time event sequence observed over an interval $[0, T]$:
    \begin{equation}
        S = \{(t_1, k_1), (t_2, k_2), \ldots, (t_N, k_N)\},
        \label{eq:sequence}
    \end{equation}
    where $0 < t_1 < t_2 < \cdots < t_N < T$, and $t_i$ denotes the timestamp of the $i$-th event. The event type, $k_i \in K = \{1, \ldots, K\}$, represents the categorical or textual type of the event. The event history prior to time $t$ is defined as $H_t = \{(t_j, k_j) \mid t_j < t\}$, which serves as the context for predicting future events.
        
    In temporal point processes, the conditional intensity function $\lambda(t, k \mid H_t)$ represents the instantaneous rate at which an event of type $k$ occurs at time $t$, conditioned on the historical context $H_t$. This function is given by:  
    \begin{equation}
            \lambda(t, k \mid H_t) =
            \lim_{\Delta t \to 0^+}
            \frac{P(k \text{ occurs in } [t, t+\Delta t) \mid H_t)}{\Delta t}.
            \label{eq:intensity}
    \end{equation}

    This intensity function is fundamental for governing the stochastic dynamics of event sequences in continuous time. It quantifies how likely an event is to occur at any given time, based on the prior event history. The log-likelihood of an observed sequence $S$ is formulated as:
    \begin{equation}
            \mathcal{L}_{\text{TPP}} =
            \sum_{i=1}^{N} \log \lambda(t_i, k_i \mid H_{t_i})
            - \int_0^T \sum_{k \in K} \lambda(t, k \mid H_t)\, dt,
            \label{eq:likelihood}
    \end{equation}
    where the first term maximizes the likelihood of the observed events, ensuring the model fits the actual occurrences. The second term enforces the probability of no events occurring during inactive intervals, acting as a regularizer. This ensures the model accounts for periods where no events take place.

    Given a learned intensity function $\lambda(t)$ for each event type $k$, the next event's time and type can be estimated under the \textit{Minimum Bayes Risk (MBR)} principle \cite{bertsch2023s}. This principle minimizes the expected prediction error and is formulated as:
        \begin{equation}
            \begin{aligned}
                \hat{t}_i &=
                \int_{t_{i-1}}^{\infty}
                t \, \lambda(t)
                \exp\!\left(
                    -\int_{t_{i-1}}^{t}
                    \lambda(s) \, ds
                \right) dt, \\
                &\qquad \lambda(t) = \sum_{k \in K} \lambda_k(t).
            \end{aligned}
        \end{equation}
    
Here, $\hat{t}_i$ represents the predicted time of the next event, computed by integrating the intensity function $\lambda(t)$ over time, while accounting for the decay factor, which represents the diminishing influence of past events over time. The total intensity function $\lambda(t)$ is the sum of the intensity functions for all event types, reflecting the combined effect of all possible event occurrences.
        
To predict the type of the next event $\hat{k}_i$, we select the event type $k$ that maximizes the intensity function at the predicted time:
     \begin{equation}
            \hat{k}_i = \arg\max_{k \in K} \lambda_k(\hat{t}_i).
     \end{equation}

The expectation in $\hat{t}_i$ can be efficiently approximated via Monte Carlo integration \cite{zuo2020transformer}, enabling probabilistic estimation of the next event's time and type. Such a method allows for accurate predictions even in the presence of noisy or sparse data by accounting for the uncertainty in continuous-time modeling.

\subsection{Overall Framework}

While neural TPPs parameterize the conditional intensity function $\lambda(t, k \mid H_t)$ using recurrent or attention-based architectures \cite{du2016recurrent, mei2017neural, zuo2020transformer}, they typically overlook the rich semantic information embedded in textual and multimodal data. This limitation hinders their ability to jointly model temporal dynamics and contextual meaning in continuous-time event sequences.

\begin{figure*}
    \centering
    \includegraphics[trim=0 0 0 0,clip,scale=0.4]{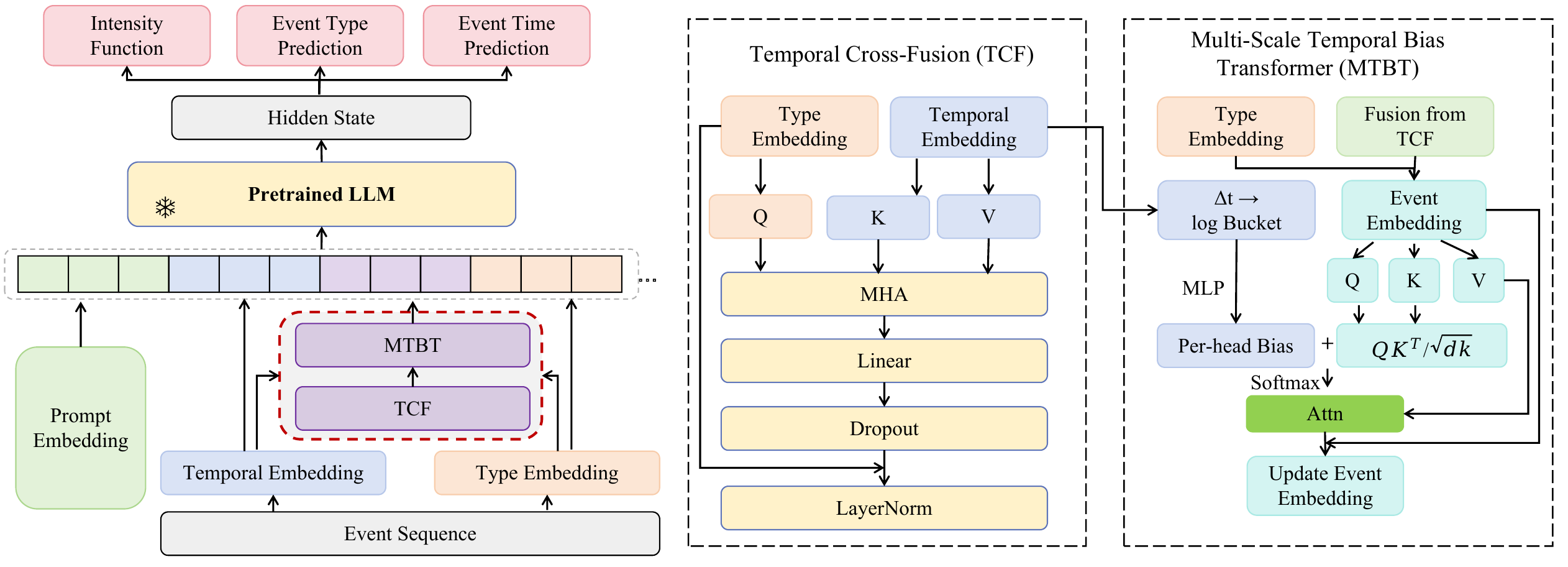}
    \caption{The TPP-TAL framework for continuous-time event modeling. Given an input event sequence, each event is encoded with both temporal and type embeddings, which are jointly processed through two plug-and-play modules: Temporal Cross-Fusion (TCF) and Multi-Scale Temporal Bias Transformer (MTBT). TCF performs fine-grained fusion between temporal and semantic features within individual events, while MTBT captures multi-scale temporal dependencies across events through per-head temporal biases. The fused representations are then fed into a pretrained LLM, which generates contextual hidden states for downstream tasks, including event type prediction, event time prediction, and intensity function estimation.}
    \label{fig:framework}
\end{figure*}

To address this, we propose TPP-TAL, a framework designed to enhance temporal awareness in large language models. The core innovation is the explicit integration of temporal signals into the semantic reasoning process, enabling the model to capture both \textit{what happens} and \textit{when it happens} through unified temporal-semantic representation learning. As shown in Fig. \ref{fig:framework}, TPP-TAL introduces two complementary plug-and-play modules: the Temporal Cross-Fusion (TCF) module and the Multi-Scale Temporal Bias Transformer (MTBT). The TCF module performs fine-grained fusion of temporal and semantic embeddings within individual events using cross-attention, enabling dynamic, time-conditioned semantic representations. MTBT extends this by introducing learnable time-dependent biases into the attention mechanism, capturing temporal dependencies across events at multiple scales, including short-term triggers and long-range periodicity. These fused temporal-semantic representations are then interleaved and fed into a pretrained LLM, without modifying its parameters. This design preserves the LLM’s strong semantic reasoning capabilities while adding explicit temporal awareness. The resulting temporally informed hidden states can be utilized for a variety of subsequent tasks, including event classification, time forecasting, and intensity assessment.

\subsection{Event Representation}
    
In a temporal event sequence, each event has two fundamental components: semantic information and temporal characteristics. The semantic component represents the meaning or categorical type of the event, while the temporal component captures when it occurs and how it evolves over time. To jointly model these complementary aspects, each event $(t_i, k_i)$ is encoded using both semantic and temporal embeddings, which reside in a shared latent space.

The semantic embedding captures the contextual and categorical meaning of each event. The event type $k_i$ is represented as a sequence of tokens processed by the language model. Let $x_i$ = \{$x_{i,1}$, $x_{i,2}$, $\ldots$, $x_{i,L_i}\}$ denote the token sequence corresponding to event type $k_i$, where $L_i$ is the number of tokens after tokenization. The token embeddings are obtained through a learnable embedding layer:
     \begin{equation}
        X_i = [x_{i,1}, x_{i,2}, \ldots, x_{i,L_i}] \in \mathbb{R}^{L_i \times D}.
    \end{equation}

This preserves fine-grained semantic representations within each event and allows the model to capture contextual dependencies among tokens. For discrete event types without textual descriptions, this formulation naturally simplifies to a single learnable embedding vector, $e_k(k_i)$ = $W_k[k_i]$, where $W_k \in \mathbb{R}^{|K| \times D}$.
    
The temporal embedding encodes the timing characteristics of events. Each timestamp $t_i$ is projected into a continuous embedding vector through a learnable mapping function $f_t(\cdot)$:
    \begin{equation}
        e_t(t_i) = f_t(t_i).
    \end{equation}

The function $f_t(\cdot)$ can take various forms depending on the temporal nature of the data. A linear projection models monotonic time progression, sinusoidal encoding captures periodic behaviors, and a learnable nonlinear transformation based on time intervals enables flexible modeling of irregular temporal dependencies. Both semantic and temporal embeddings are aligned in the same dimensional space, providing a unified foundation for subsequent time–semantic fusion and reasoning.

\subsection{Temporal Cross-Fusion (TCF)}

The TCF module integrates temporal information into event semantics in a unified and learnable way. In continuous-time event modeling, the significance of semantic tokens (e.g., textual descriptions or categorical embeddings) often depends on the timing of events—events occurring in close succession tend to reinforce each other's meaning, while those separated by longer intervals exhibit weaker contextual relevance. To model this time-dependent semantic modulation, TCF offers a unified framework for combining temporal and semantic embeddings through one of three strategies: additive fusion, concatenation fusion, or cross-attention fusion. Among these, cross-attention fusion is adopted as the default due to its superior expressiveness and flexibility, enabling more dynamic and context-sensitive interaction between temporal and semantic components.

Let the token matrix of the $i$-th event type embedding be $X_i$ = $[x_{i,1}, x_{i,2}, \ldots, x_{i,L_i}] \in \mathbb{R}^{L_i \times D}$, and its temporal embedding be $e_t(t_i) \in \mathbb{R}^{D}$. TCF outputs a fused representation $\tilde{X}_i \in \mathbb{R}^{L_i \times D}$ that combines both semantic and temporal information, thereby enriching the event representation with context-aware temporal features.
    
\textbf{Additive fusion}. The simplest strategy projects the time embedding into the semantic space and adds it to all tokens:
    \begin{equation}
        \tilde X_i = \mathrm{LN}\!\big(X_i + \mathrm{Dropout}(W_t\,e_t(t_i)\,\mathbf{1}_{L_i}^\top)\big),
    \end{equation}
    where $W_t \in \mathbb{R}^{D\times D}$ is a learnable projection matrix, $\mathbf{1}_{L_i}$ is a vector of ones, and $\mathrm{LN}$ denotes layer normalization. This method is similar to feature-wise linear modulation (FiLM) \cite{perez2018film}, applying a uniform temporal shift across all tokens. While computationally efficient, it does not capture token-level temporal effects, making it less expressive when detailed temporal context is crucial.

    \textbf{Concatenation fusion}. To capture richer temporal-semantic interactions, the time embedding is concatenated with each token and projected back to the original dimension:
    \begin{equation}
    \tilde X_i = \mathrm{LN}\!\big(X_i + \mathrm{Dropout}([X_i \,\Vert\, e_t(t_i)\,\mathbf{1}_{L_i}^\top]\,W_c)\big),
    \end{equation}
    where $W_c \in \mathbb{R}^{2D\times D}$ is trainable, and $\Vert$ denotes concatenation across the feature dimension. This formulation enables nonlinear fusion of time and semantics while still applying the same temporal influence to all tokens. While it provides a richer interaction than additive fusion, it still lacks the flexibility to dynamically adjust temporal influences at the token level.

    \textbf{Cross-attention fusion}. To overcome these limitations, TCF employs a cross-attention mechanism, where the temporal embedding serves as a learnable \textit{key–value} context modulating each semantic token query:
    \begin{equation}
    Q = X_i\,W_Q,\quad K = e_t(t_i)\,W_K,\quad V = e_t(t_i)\,W_V,
    \end{equation}
    where $W_Q, W_K, W_V \in \mathbb{R}^{D\times D}$ are learnable projection matrices. After splitting into $H$ heads, we obtain $Q \in \mathbb{R}^{H \times L_i \times d_k}$ and $K, V \in \mathbb{R}^{H \times 1 \times d_k}$. The attention operation is computed as:
    \begin{equation}
    \mathrm{Attn}(Q,K,V) = \mathrm{softmax}\!\left(\frac{Q\,K^\top}{\sqrt{d_k}}\right)V.
    \end{equation}
    The fused output is then obtained as:
    \begin{equation}
    \tilde X_i = \mathrm{LN}\!\big(X_i + \mathrm{Dropout}(\mathrm{Attn}(Q,K,V)\,W_O)\big),
    \end{equation}
    where $W_O \in \mathbb{R}^{D\times D}$ is the output projection. This formulation preserves the structure and efficiency of standard Transformer blocks while enabling dynamic, time-conditioned semantic modulation. The cross-attention mechanism allows the model to effectively modulate each token’s representation based on the event’s temporal context, providing a more fine-grained and context-sensitive integration of temporal and semantic information.

    Finally, the event-level representation is computed by mean pooling:
        \begin{equation}
        s_i = \frac{1}{L_i} \sum_{l=1}^{L_i} \tilde x_{i,l}.
        \end{equation}
    The resulting vector $s_i$ captures both the semantic content and temporal context of each event. Since TCF operates before the LLM input layer, it introduces temporal awareness in a lightweight, modular, and plug-and-play manner without altering pretrained parameters.

\subsection{Multi-Scale Temporal Bias Transformer (MTBT)}

While the TCF module focuses on local temporal-semantic fusion within individual events, dependencies between events often unfold across multiple time scales, ranging from immediate triggers to long-term periodic or structural patterns. To capture these hierarchical temporal relationships, we propose the Multi-Scale Temporal Bias Transformer (MTBT), which augments the standard self-attention mechanism with learnable temporal bias terms that explicitly depend on the time intervals between events. This modification enables the model to regulate inter-event attention weights based on temporal distance, embedding temporal awareness directly into the Transformer architecture and allowing the model to better capture complex long-term dependencies and event dynamics.
    
Each event representation $s_i$ obtained from the TCF module is linearly projected into query, key, and value representations as $Q_i$ = $s_i W_Q$, $K_i$ = $s_i W_K$, and $V_i$ = $s_i W_V$, where $W_Q$, $W_K$, and $W_V \in \mathbb{R}^{D \times D}$ are learnable projection matrices. Given two events $i$ and $j$ occurring at timestamps $t_i$ and $t_j$, respectively, their temporal distance is defined as $\Delta t_{ij}$ = $|t_i - t_j|$. For each attention head $h$, the attention score is computed as:

    \begin{equation}
    \text{score}_{ij}^{(h)} = 
    \frac{Q_i^{(h)} (K_j^{(h)})^\top}{\sqrt{d_k}} + b_{ij}^{(h)},
    \label{eq:mtbt_score}
    \end{equation}
    where $b_{ij}^{(h)}$ is a learnable temporal bias that modulates the attention strength based on the time gap $\Delta$ $t_{ij}$. Intuitively, this bias enables the model to assign greater attention to temporally proximate events while gradually diminishing the influence of distant events. As a result, MTBT learns to emphasize temporally relevant context and naturally incorporates temporal decay into its attention mechanism.

    However, time intervals in real-world sequences can vary significantly, which may cause numerical instability if modeled directly. To address this, we adopt a logarithmic bucketization strategy \cite{dhingra2022time}, discretizing time intervals into $B$ logarithmically spaced bins:
    \begin{equation}
    b_{ij} =
    \left\lfloor
    (B - 1)\,
    \frac{
    \log(\Delta t_{ij} + \epsilon) - \log(\Delta t_{\min} + \epsilon)
    }{
    \log(\Delta t_{\max} + \epsilon) - \log(\Delta t_{\min} + \epsilon)
    }
    \right\rfloor,
    \label{eq:bucketization}
    \end{equation}
    where $\epsilon$ prevents numerical overflow. Each discrete bucket index $b_{ij}$ is associated with a learnable embedding $z_{ij} = \mathrm{Emb}(b_{ij}) \in \mathbb{R}^{d_b}$, which is further transformed by a multilayer perceptron (MLP) to produce the temporal bias vector:  
    \begin{equation}
        b_{ij} = \mathrm{MLP}(z_{ij}) \in \mathbb{R}^{H}.
        \label{eq:bias_mlp}
    \end{equation}

    This strategy compresses large temporal ranges while preserving fine-grained short-term dynamics, allowing the model to learn both local and global temporal correlations in a numerically stable manner, and ensuring robust performance across varying time scales.

    To enhance the expressiveness of temporal modeling, MTBT introduces a per-head temporal bias mechanism \cite{daniel2024ddlp}:
    \begin{equation}
    b_{ij}^{(h)} = b_{ij}[h],
    \label{eq:per_head_bias}
    \end{equation}
    which allows each attention head to focus on distinct temporal scales, with some heads concentrating on short-term activation trends, and others capturing long-term dependencies or periodic cycles. This design naturally enables a multi-scale temporal decomposition, where different heads collaboratively model the diverse rhythms of temporal evolution, providing a more comprehensive representation of event dynamics.

    Given the temporal biases, the attention weights become explicitly time-aware:
    \begin{equation}
        \alpha_{ij}^{(h)} = \mathrm{softmax}_j(\text{score}_{ij}^{(h)}), 
    \end{equation}
    \begin{equation}
        y_i^{(h)} = \sum_j \alpha_{ij}^{(h)} V_j^{(h)}.
        \label{eq:attention_update}
    \end{equation}
   Outputs from all heads are concatenated and passed through a feed-forward layer:
    \begin{equation}
        y_i = \mathrm{Concat}(y_i^{(1)}, \ldots, y_i^{(H)}) W_O, 
    \end{equation}
    \begin{equation}
        s_i' = \mathrm{FFN}(\mathrm{LN}(s_i + y_i)).
        \label{eq:mtbt_output}
    \end{equation}

    This architecture preserves the efficiency and compatibility of the Transformer while integrating explicit temporal reasoning into its attention mechanism. By modeling temporal dependencies across multiple scales, MTBT effectively captures both short-term triggers and long-range temporal structures, enhancing the model’s ability to handle complex temporal patterns in real-world data.

\subsection{Integration with the LLM Input}

    After the TCF module performs fine-grained temporal-semantic fusion within individual events and the MTBT module captures multi-scale temporal dependencies across events, the model produces a sequence of enhanced event representations \{$s'_1$, $s'_2$, $\ldots$, $s'_N$\}. Each event $(t_i, k_i)$ is represented by its temporal embedding $e_t(t_i)$, semantic token matrix $X_i$, and the MTBT-enhanced vector $s'_i$, which jointly encode both local semantics and global temporal context. Following the input structure of TPP-LLM \cite{liu2025tpp}, a learnable prompt embedding $P$ is prepended to the event sequence, serving as a global context initializer for the LLM. The final hybrid input sequence is then constructed as $X_{\text{LLM}}$ = [$P$, $e_t(t_1)$, $X_1$, $s'_1$, $e_t(t_2)$, $X_2$, $s'_2$, $\ldots$, $e_t(t_N)$, $X_N$, $s'_N$], where $N$ denotes the total number of events. This sequence is padded to a uniform length and directly fed into the pretrained LLM. The model produces contextual hidden states $H = \mathrm{LLM}(X_{\text{LLM}})$, where $H \in \mathbb{R}^{L \times D}$ represents the hidden states for all token positions.
        
    For each event $i$, the hidden state $h_i$ corresponding to the final token in its segment is extracted as the contextualized event representation. This representation serves as the foundation for downstream tasks such as event type classification, timestamp regression, and intensity estimation. By incorporating temporally enriched embeddings into the LLM input, the framework enables unified reasoning over both temporal and semantic dimensions. Importantly, this is achieved without altering pretrained parameters, preserving the architectural compatibility and computational efficiency of the LLM, while providing a cohesive foundation for continuous-time event modeling. The integration of temporal awareness directly into the LLM input enables the model to utilize its strong language understanding capabilities while being sensitive to time-dependent dynamics in event sequences.

\subsection{Training Objective}

The model is trained with a unified multi-task objective that jointly captures both temporal dynamics and semantic understanding. The overall loss is formulated as:
    \begin{equation}
        \mathcal{L} =
        - \mathcal{L}_{\text{TPP}} 
        + \alpha \mathcal{L}_{\text{type}} 
        + \beta \mathcal{L}_{\text{time}},
    \end{equation}
    where - $\mathcal{L}_{\text{TPP}}$ represents the negative log-likelihood derived from the temporal point process, encouraging the model to precisely model the probabilistic behavior of event occurrences over continuous time. $\mathcal{L}_{\text{type}}$ is the cross-entropy loss for event type classification, which guides the model to correctly predict the semantic category of future events. $\mathcal{L}_{\text{time}}$ is the mean-squared error loss for timestamp regression, ensuring accurate estimation of event occurrence times. The hyperparameters $\alpha$ and $\beta$ control the relative importance of each task. This joint optimization enables the model to capture both temporal and semantic dependencies in a unified framework, facilitating coherent and robust modeling of continuous-time event sequences. By learning to balance these objectives, the model can better integrate temporal awareness and semantic reasoning, resulting in improved event prediction performance.

\section{Experimental Results} \label{Experiment}
\subsection{Experimental Setup}

We conduct extensive experiments on four representative continuous-time event sequence datasets covering natural, social, and commercial domains. These datasets comprehensively evaluate the proposed model's capability to capture temporal dependencies and semantic intricacies across diverse real-world scenarios, with their characteristics summarized in Table \ref{table:datasets_characteristics}.

    \begin{itemize}
        \item US-EQ (United States Earthquake) \footnote{The dataset is provided by the China Earthquake Data Center (\href{http://data.earthquake.cn} {http://data.earthquake.cn}).}:  
        This dataset records earthquake occurrences across the United States over more than a decade, including precise timestamps, geographic coordinates, and magnitudes. 
    
        \item SOF (Stack Overflow) \cite{jure2014snap}:  
        Derived from Stack Overflow’s question–answer interactions, this dataset reflects dynamic user engagement patterns through timestamps, topic categories, and relational metadata. 
    
        \item AMZ (Amazon Review) \cite{ni2019justifying}:  
        This dataset consists of Amazon product review activities, including review timestamps, product categories, and sentiment polarity annotations. 
    
        \item NYC (New York City Taxi) \cite{whong2014foiling}:  
        This dataset captures taxi pick-up and drop-off events in New York City, including trip start and end times as well as geographic coordinates.  
    \end{itemize}

    \begin{table*}[htbp]
        \centering
        \caption{Characteristics of different datasets}
        \begin{tabular}{lccccl}
        \toprule
        \textbf{Dataset} & \textbf{Event Types} & \textbf{Total Events} & \textbf{Sequences} & \textbf{Avg Sequence Length} & \textbf{Characteristics} \\
        \midrule
        \textbf{US-EQ} & 3 & 29,521 & 3,009 & 9.81/day & Sparse events, long time span \\
        \textbf{SOF} & 25 & 187,836 & 3,336 & 56.31/month & High-frequency events\\
        \textbf{AMZ} & 18 & 127,054 & 2,245 & 56.59/week & Moderate density, random review times \\
        \textbf{NYC} & 8 & 362,374 & 2,957 & 122.55/hour & High-frequency, time periodicity\\
        \bottomrule
        \end{tabular}
        \label{table:datasets_characteristics}
    \end{table*}

We use TPP-LLM \cite{liu2025tpp} as the baseline model for comparison, mainly due to the limited availability of large-scale pretrained models specifically designed for temporal point processes. Although some recent studies have explored integrating neural sequence models with temporal dynamics, many of these approaches are either unpublished or only partially open-sourced, making comprehensive and reproducible comparisons difficult. TPP-LLM is one of the most well-established models in this field. By combining a large language model with temporal point process modeling, TPP-LLM provides a unified framework for reasoning over both semantic context and temporal dynamics. Its strong performance in event type classification and time prediction tasks makes it a suitable baseline for evaluating our proposed model. The pretrained model used in our experiments is TinyLlama-1.1B-Chat-v1.0, a lightweight variant of Llama optimized for chat-based tasks. By optimizing both computational efficiency and performance, this model is well-suited for adapting to temporal point process tasks, enabling effective fine-tuning within our framework.
    
All experiments were carried out utilizing the PyTorch framework and executed on a single NVIDIA RTX 3090 GPU.  To ensure experimental fairness and reproducibility, all models are trained under identical computational environments with fixed random seeds.  To balance memory usage and training stability, the batch size is adjusted according to dataset scale: 6 for SOF, 2 for NYC, 8 for US-EQ, and 2 for AMZ.  Furthermore, the maximum number of training epochs is set to 20, and the learning rate is fixed at \(1 \times 10^{-5}\), a value empirically verified to be stable for large language model fine-tuning.  The loss weighting coefficients, \(\beta_{\text{type}}\) and \(\beta_{\text{time}}\), are both set to 1 to maintain balanced optimization between event type classification and time prediction objectives. Additionally, the number of integral samples is fixed at 20, providing a reasonable trade-off between computational efficiency and numerical accuracy in temporal integration. The bucket size $B$ is set to 32, which allows for an effective discretization of time intervals and helps capture both short-term and long-range temporal relationships. For consistency, the prompt configuration strictly follows the original TPP-LLM setup, ensuring alignment in input representation and contextual conditioning.  Except for minor adjustments in batch size, all hyperparameters are kept consistent across datasets to ensure fairness and stability throughout the training process.

\subsection{Evaluation Metrics}

In order to thoroughly assess the effectiveness of the proposed model in continuous-time event modeling, we employ three complementary metrics: Log-Likelihood (\textit{LL}), Accuracy (\textit{Acc}), and Root Mean Squared Error (\textit{RMSE}). These metrics assess the model’s performance in event generation, semantic classification, and temporal prediction, providing a well-rounded view of its effectiveness and robustness in time-semantic fusion.
    
The \textit{LL}, as defined in Equation \ref{eq:likelihood}, measures how well the model fits the observed event sequences in continuous time by modeling the conditional intensity function of the temporal point process. As a core metric in TPP-based modeling, \textit{LL} reflects the model’s ability to reproduce the probabilistic structure and dynamic evolution of event sequences. A higher log-likelihood indicates better modeling of temporal patterns and event dependencies, showing stronger global temporal reasoning and generative consistency.

    \textit{Acc} evaluates the model’s semantic understanding and discriminative ability in event-type prediction. It is computed as:
    \begin{equation}
        \textit{Acc} = \frac{N_{\text{correct}}}{N_{\text{total}}},
    \end{equation}
    where $N_{\text{correct}}$ refers to the count of events accurately predicted, while $N_{\text{total}}$ represents the complete set of events considered. A higher accuracy reflects better semantic representation learning and contextual reasoning across heterogeneous event types.

    \textit{RMSE} quantifies the model’s precision in temporal prediction by calculating the average deviation between the predicted and true event times:
    \begin{equation}
        \textit{RMSE} = \sqrt{\frac{1}{N}\sum_{i=1}^{N}(t^{\text{pred}}_i - t^{\text{true}}_i)^2},
    \end{equation}
    where $t^{\text{pred}}_i$ and $t^{\text{true}}_i$ represent the predicted and ground-truth timestamps of the $i$-th event.  \textit{RMSE} reveals how effectively the model captures the timing regularities and dynamic rhythms of events. Lower \textit{RMSE} values indicate higher temporal precision and finer-grained modeling of time intervals.

\subsection{Results and Analysis}

A comprehensive comparison is conducted between the baseline TPP-LLM, the proposed TPP-TAL model, and two ablated variants derived from the proposed architecture. In the w/o MTBT configuration, the Multi-Scale Temporal Bias Transformer module is removed while retaining the Temporal Cross-Fusion mechanism to assess the contribution of multi-scale temporal modeling. In the w/o TCF variant, the Temporal Cross-Fusion module is removed, leaving only the cross-event temporal dependency modeling to evaluate the influence of local time-semantic interactions on performance.

As shown in Table \ref{tab:ll}, TPP-TAL achieves the highest or near-highest log-likelihood scores across most datasets, with particularly pronounced improvements on the US-EQ dataset. These results confirm that the proposed framework effectively captures the complex temporal dynamics and latent dependency structures underlying event-time distributions. For example, on US-EQ, which involves sparse events over a long time span, the model benefits significantly from the joint integration of local temporal-semantic fusion and multi-scale temporal dependencies, as seen from the marked improvement in log-likelihood scores. In contrast, when only the TCF module is used, the model shows a reduced ability to capture these dependencies, reflecting that while local temporal-semantic fusion improves intra-event coherence, it cannot fully capture the broader temporal dependencies across events. On the other hand, the MTBT module enhances global temporal reasoning by explicitly capturing long-range dependencies and multi-scale temporal patterns through learnable temporal biases. This improvement is particularly significant in datasets with longer event intervals. When TCF and MTBT are combined in TPP-TAL, the model can effectively balance local and global temporal reasoning, leading to more stable and precise event-time likelihood modeling.

    \begin{table}[h!]
    \centering
    \caption{Log-Likelihood evaluation across four datasets.}
    \label{tab:ll}
    \begin{tabular}{lcccc}
    \hline
    \textbf{Model} & \textbf{SOF} & \textbf{NYC} & \textbf{US-EQ} & \textbf{AMZ} \\
    \hline
    TPP-LLM & -1.8482 & 0.2129 & -0.4675 & -1.0470 \\
    w/o MTBT & -1.8471 & 0.2325 & -0.4663 & -1.0166 \\
    w/o TCF & -0.7108 & \textbf{0.4577} & 0.0420 & \textbf{-0.4360} \\
    TPP-TAL & \textbf{-0.4748} & 0.4477 & \textbf{0.2247} & -0.6069 \\
    \hline
    \end{tabular}
    \end{table}

For accuracy (Table \ref{tab:acc}), TPP-TAL exhibits significant improvements in semantically complex or temporally irregular datasets, such as SOF. This highlights the effectiveness of its joint temporal-semantic representation in event type prediction. The TCF module enhances semantic learning by adaptively modulating semantic token representations with temporal signals through cross-attention, enabling the model to distinguish between temporally close yet semantically diverse events. This is especially useful in datasets where the event types are diverse and the timing is irregular, allowing TPP-TAL to maintain strong predictive performance. In contrast, for high-frequency datasets like NYC, where event intervals are short and contextual variability is limited, the MTBT module alone yields only a slight accuracy improvement. This indicates that under such conditions, global temporal regularities dominate, and fine-grained local temporal-semantic interactions play a smaller role. The model performs best when both TCF and MTBT are used together, as they complement each other by combining fine-grained local sensitivity with broader temporal coherence.

 \begin{table}[h!]
    \centering
    \caption{Accuracy evaluation across four datasets.}
    \label{tab:acc}
    \begin{tabular}{lcccc}
    \hline
    \textbf{Model} & \textbf{SOF} & \textbf{NYC} & \textbf{US-EQ} & \textbf{AMZ} \\
    \hline
    TPP-LLM & 0.4386 & 0.9136 & 0.6290 & 0.6902 \\
    w/o MTBT & 0.4401 & 0.9170 & 0.6353 & 0.6939 \\
    w/o TCF & 0.7288 & \textbf{0.9612} & \textbf{0.8482} & \textbf{0.8112} \\
    TPP-TAL & \textbf{0.7962} & 0.9181 & 0.8430 & 0.7502 \\
    \hline
    \end{tabular}
    \end{table}

With respect to RMSE (Table \ref{tab:rmse}), TPP-TAL consistently achieves the lowest prediction errors across all datasets, clearly outperforming both the baseline and single-module variants. This demonstrates its superior precision and stability in temporal prediction tasks. When the model includes only the MTBT or TCF module, performance improves modestly but remains limited. However, when both modules are integrated, the error decreases substantially, showing the strong complementary effect of the two modules. The MTBT module captures long-term dependencies and periodic patterns across events through multi-scale temporal biases, providing global constraints for temporal dynamics. On the other hand, the TCF module refines intra-event temporal-semantic fusion, enabling the model to respond better to fine-grained local variations. The combination of these two modules endows the model with both global temporal awareness and local adaptability, leading to more reliable time predictions and stronger generalization in complex continuous-time event modeling scenarios.
 
    \begin{table}[h!]
    \centering
    \caption{RMSE evaluation across four datasets.}
    \label{tab:rmse}
    \begin{tabular}{lcccc}
    \hline
    \textbf{Model} & \textbf{SOF} & \textbf{NYC} & \textbf{US-EQ} & \textbf{AMZ} \\
    \hline
    TPP-LLM & 0.4819 & 0.8920 & 0.2850 & 0.5910 \\
    w/o MTBT & 0.4933 & 0.8881 & 0.2788 & 0.5823 \\
    w/o TCF & 0.4659 & 0.8323 & 0.2878 & 0.5934 \\
    TPP-TAL & \textbf{0.3641} & \textbf{0.4426} & \textbf{0.2258} &\textbf{0.4236} \\
    \hline
    \end{tabular}
    \end{table}

\subsection{Ablation Study}

To evaluate the impact of each module design on overall model performance, we conducted a systematic ablation study focusing on two key components: the fusion strategies within the TCF module (Table \ref{table: tcf}) and the main architectural elements of the MTBT module (Table \ref{table: MTBT}).

    \begin{table}[h!]
    \centering
    \caption{Ablation results of different TCF fusion modes. }
    \label{table: tcf}
    \begin{tabular}{l l c c c}
    \hline
    \textbf{Dataset} & \textbf{Variant} & \textbf{\textit{LL}} & \textbf{\textit{Acc}} & \textbf{\textit{RMSE}} \\
    \hline
    \multirow{4}{*}{US-EQ} 
     & None & -0.4675 & 0.6290 & 0.2850 \\
     & Additive & -0.4630 & 0.6294 & 0.2787 \\
     & Concatenation & -0.4701 & 0.6305 & 0.2850 \\
     & Cross-Attention & \textbf{-0.4663} & \textbf{0.6353} & \textbf{0.2788} \\
    \hline  
    \multirow{4}{*}{SOF} 
     & None & -1.8482 & 0.4386 & 0.4819 \\
     & Additive & -1.8598 & 0.4396 & 0.5073 \\
     & Concatenation & \textbf{-1.8462} & 0.4390 & 0.4858 \\
     & Cross-Attention & -1.8471 & \textbf{0.4401} & \textbf{0.4933} \\
    \hline
    \multirow{4}{*}{AMZ} 
     & None & -1.0382 & 0.6924 & 0.5983 \\
     & Additive & -1.0233 & 0.6932 & 0.5877 \\
     & Concatenation & \textbf{-1.0202} & \textbf{0.6944} & \textbf{0.5800} \\
     & Cross-Attention & -1.0166 & 0.6939 & 0.5823 \\
    \hline
    \multirow{4}{*}{NYC} 
     & None & 0.2055 & 0.9140 & 0.9065 \\
     & Additive & \textbf{0.2399} & 0.9167 & \textbf{0.8817} \\
     & Concatenation & 0.2255 & 0.9164 & 0.9178 \\
     & Cross-Attention & 0.2325 & \textbf{0.9170} & 0.8881 \\
    \hline
    \end{tabular}
    \end{table}

As shown in Table \ref{table: tcf}, the four temporal and semantic fusion methods (where None denotes pure text input without temporal information) show distinct performance differences across datasets. Among these strategies, the Cross-Attention fusion method consistently achieves the best results across most datasets, demonstrating strong robustness and generalization. The learnable cross-attention structure allows time embeddings to dynamically influence semantic tokens, enabling the model to capture fine-grained temporal dependencies between asynchronous events while maintaining semantic consistency. On datasets such as SOF and US-EQ, this approach significantly improves temporal sensitivity and semantic alignment, showing its effectiveness in handling complex and irregular event sequences. In contrast, the Additive fusion method provides only modest gains, particularly on datasets with short intervals and highly regular event patterns, such as NYC. Its linear modulation limits its ability to model localized semantic differences and nonlinear temporal relationships. The Concatenation method performs more consistently on datasets like AMZ, where semantic patterns are relatively stable and temporal variations are moderate. This suggests that, when the time-semantic relationship is smoother, the Concatenation method strikes a good balance between performance and computational cost.

Overall, the Cross-Attention fusion method excels in complex semantic and irregular temporal settings, providing precise modeling of time-semantic interactions. Meanwhile, the Additive and Concatenation methods remain practical for high-frequency or more regular event sequences.

    \begin{table}[h!]
    \centering
    \caption{Ablation results of the MTBT module. }
    \label{table: MTBT}
    \begin{tabular}{l l c c c}
    \hline
    \textbf{Dataset} & \textbf{Variant} & \textbf{\textit{LL}} & \textbf{\textit{Acc}} & \textbf{\textit{RMSE}} \\
    \hline
    \multirow{4}{*}{US-EQ} 
     & MTBT & \textbf{0.0420} & \textbf{0.8482} & 0.2878 \\
     & w/o Bias & -0.2306 & 0.7431 & 0.2974 \\
     & w/o Log bucket & -0.0569 & 0.8182 & 0.2934 \\
     & Shared  bias & -0.0773 & 0.8145 & \textbf{0.2881} \\
    \hline
    \multirow{4}{*}{SOF} 
     & MTBT & -0.7108 & \textbf{0.7288} & \textbf{0.4659} \\
     & w/o Bias & -1.6544 & 0.4598 & 0.5015 \\
     & w/o Log bucket & -0.8197 & 0.6758 & 0.4656 \\
     & Shared  bias & \textbf{-0.7033} & 0.7294 & 0.4713 \\
    \hline
    \multirow{4}{*}{AMZ} 
     & MTBT & -0.4360 & \textbf{0.8112} & 0.5934 \\
     & w/o Bias & -0.9157 & 0.7049 & 0.6175 \\
     & w/o Log bucket & -0.5460 & 0.7755 & \textbf{0.5905} \\
     & Shared  bias & \textbf{-0.4125} & 0.8103 & 0.6044 \\
    \hline
    \multirow{4}{*}{NYC} 
     & MTBT & 0.4577 & 0.9612 & 0.8323 \\
     & w/o Bias & 0.2630 & 0.9226 & 0.8633 \\
     & w/o Log bucket & 0.4578 & 0.9480 & \textbf{0.8006} \\
     & Shared  bias & \textbf{0.5530} & \textbf{0.9603} & 0.8037 \\
    \hline
    \end{tabular}
    \end{table}

The impact of removing key components from the MTBT module was evaluated by analyzing three variants: removing the temporal bias mechanism (w/o Bias), eliminating the logarithmic bucketization mechanism (w/o Log bucket), and forcing all attention heads to share the same temporal bias parameters (Shared bias). As shown in Table \ref{table: MTBT}, the complete MTBT module consistently outperforms all its variants across datasets, highlighting the critical role of the multi-scale temporal bias mechanism in capturing cross-event temporal dependencies. Specifically, in sparse and long-interval datasets such as US-EQ and SOF, removing the temporal bias results in a significant performance drop, emphasizing the importance of explicitly modeling temporal differences to represent causal temporal relationships. Without temporal bias, the model tends to misinterpret frequent co-occurrences as temporally proximate, thereby weakening its ability to capture long-range dependencies. Eliminating the logarithmic bucketization mechanism also leads to degraded performance, particularly in datasets with large temporal gaps such as US-EQ and AMZ, indicating that bucketization effectively compresses wide time intervals while maintaining high resolution for short intervals, thereby improving model stability and generalization. Furthermore, the Shared bias configuration, though it retains the bias structure, forces all attention heads to share the same parameters, limiting the model’s flexibility to model different temporal scales independently and ultimately reducing its representational power. In contrast, the complete MTBT module enables different attention heads to focus on both short-term triggers and long-term dependencies, thus facilitating hierarchical temporal modeling across complex event sequences.

\begin{figure*}[htbp]
    \centering
    
    \begin{minipage}[b]{0.9\textwidth}
        \centering
        \includegraphics[width=\textwidth]{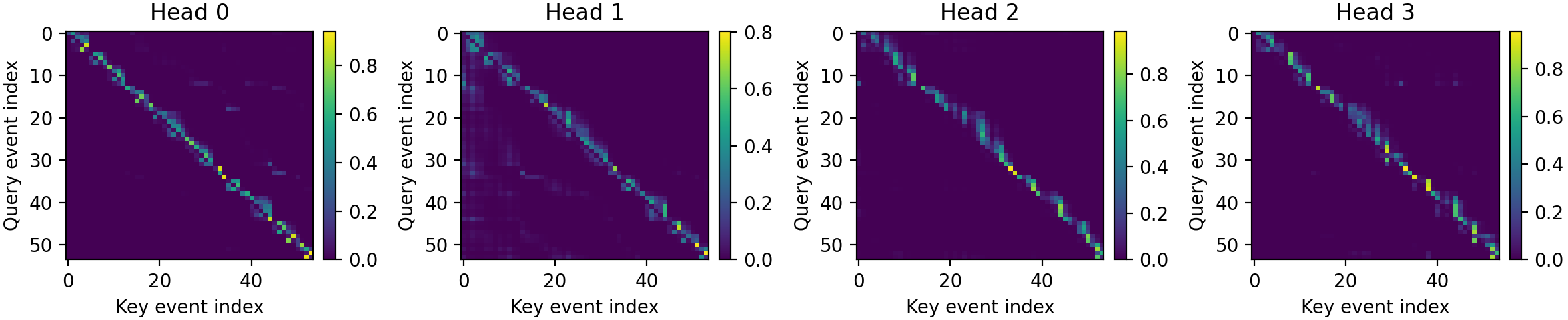}
        \vskip 0.5em
        (a) MTBT  
    \end{minipage}
    
    \vskip\baselineskip

    \begin{minipage}[b]{0.9\textwidth}
        \centering
        \includegraphics[width=\textwidth]{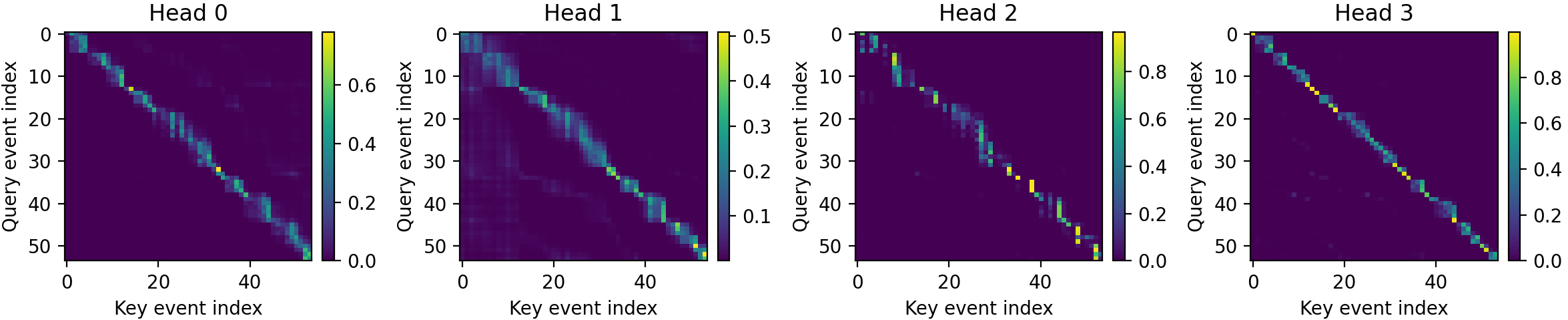}
        \vskip 0.5em
        (b) w/o Bias  
    \end{minipage}
    
    \vskip\baselineskip

    \begin{minipage}[b]{0.9\textwidth}
        \centering
        \includegraphics[width=\textwidth]{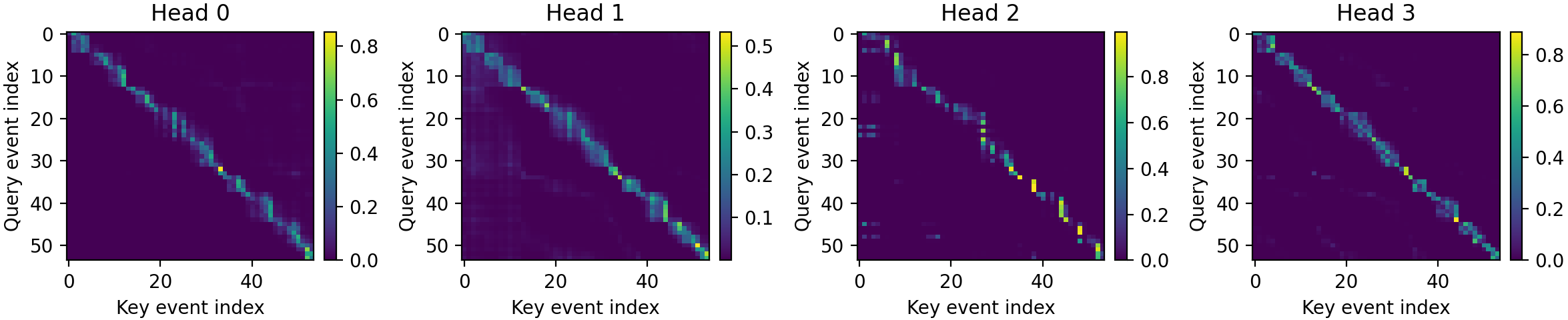}
        \vskip 0.5em
        (c) w/o Log bucket 
    \end{minipage}

    \begin{minipage}[b]{0.9\textwidth}
        \centering
        \includegraphics[width=\textwidth]{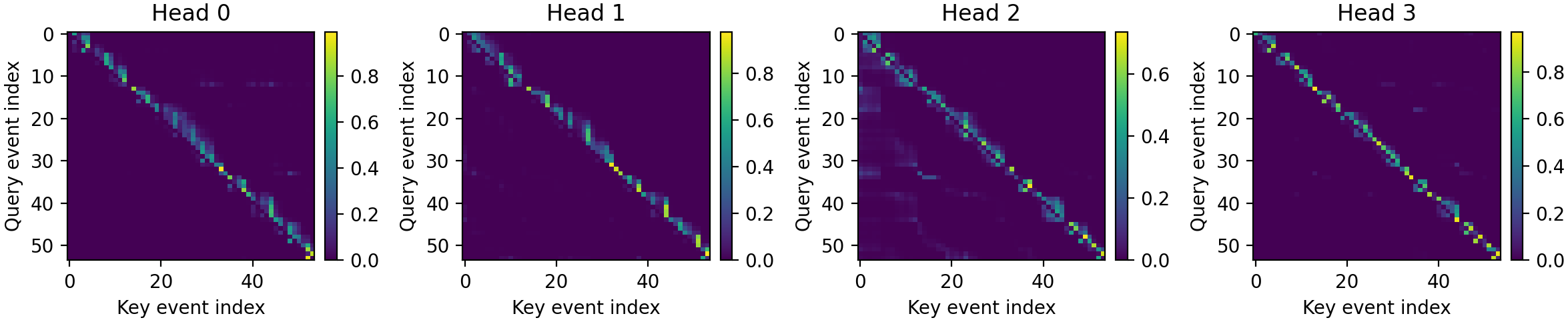}
        \vskip 0.5em
        (d) Shared bias  
    \end{minipage}    
   \caption{Visualization of Attention Distribution for Different MTBT Variants on SOF Dataset}  
    \label{fig:time_aware_heads}
\end{figure*}

Additionally, we visualized the attention patterns for different variants of the MTBT module using the SOF dataset, which contains high-frequency events. As shown in Figure \ref{fig:time_aware_heads}, the complete MTBT module exhibits a distinct attention focus along the diagonal, demonstrating the model's ability to effectively capture both short-term and long-term temporal dependencies. This is particularly crucial for high-frequency datasets like SOF, where events occur in rapid succession, and the model must distinguish and capture densely occurring events. In contrast, when the temporal bias mechanism is removed, the attention distribution becomes more uniform, suggesting that the model is less effective at identifying the fine-grained temporal dependencies inherent in high-frequency data. The variant without logarithmic bucketization results in a more scattered attention distribution, which indicates that the model struggles with managing the short time intervals between events. This becomes especially problematic when large time gaps need to be handled, as the model cannot properly adjust for small temporal variations in high-frequency settings. Lastly, when using the shared bias, the attention distribution is less focused, showing that applying a single temporal bias across different event types diminishes the model's ability to capture specific temporal patterns for individual events. These visualizations underscore the importance of both the multi-scale temporal bias and the bucketization mechanism in accurately modeling temporal dependencies in high-frequency event sequences, such as those in the SOF dataset.
    
\subsection{ Varying Number of Buckets}

To better understand the impact of different time bucket sizes on model performance, we conducted a systematic experiment using two datasets with distinct temporal characteristics: US-EQ and SOF. The results, shown in Table \ref{table: Buckets}, highlight the critical role that the number of time buckets plays in both the granularity of time discretization and the model's ability to capture and generalize temporal patterns.

On the US-EQ dataset, where events are sparse and spread over a long period, the model performs optimally with 16 buckets (\textit{LL} = 0.2724, \textit{Acc} = 0.8578, and \textit{RMSE} = 0.1827). This configuration offers a balanced approach, effectively capturing long-term temporal dependencies without overfitting to minute time variations. However, as the number of buckets increases (e.g., 32, 64, or 128), performance declines. This is due to the excessive segmentation of time, which undermines the model’s ability to capture long-range dependencies, resulting in instability and a less reliable representation of temporal dynamics. Conversely, on the SOF dataset, where events occur more frequently and time intervals are shorter, both 8 and 16 buckets yield strong performance, with 16 buckets slightly outperforming in terms of \textit{RMSE} (0.3421), while 8 buckets achieve slightly higher accuracy (\textit{Acc} = 0.8131). However, beyond 16 buckets, performance noticeably degrades, particularly in log-likelihood. This suggests that in high-frequency event scenarios, finer time segmentation leads to noise amplification, causing the model to incorrectly interpret small time differences as meaningful, thus diminishing overall predictive accuracy.

In summary, the effect of time bucket sizes on model performance is heavily influenced by the frequency and temporal structure of the dataset. For sparse datasets, such as US-EQ, a moderate number of buckets effectively captures long-range temporal dependencies, while for high-frequency datasets, such as SOF, fewer buckets are more suitable for minimizing noise interference. These results underline the importance of selecting an appropriate number of time buckets tailored to the dataset's temporal properties.

\begin{table}[h]
    \centering
    \small
    \caption{Ablation results of the number of buckets. }
    \label{table: Buckets}
    \begin{tabular}{l c c c c}
    \hline
    \textbf{Dataset} & \textbf{Number of Buckets} & \textbf{\textit{LL}} & \textbf{\textit{Acc}} & \textbf{\textit{RMSE}} \\
    \hline
    \multirow{4}{*}{US-EQ} 
     & 8 & 0.2361 & 0.8452 & 0.1829 \\
     & 16 & \textbf{0.2724} & \textbf{0.8578} & \textbf{0.1827} \\
     & 32 & 0.2247 & 0.8430 & 0.2258 \\
     & 64 & 0.2083 & 0.8386 & 0.1941 \\
     & 128 & 0.1752 & 0.8267 & 0.2022 \\
    \hline
    \multirow{4}{*}{SOF} 
     & 8 & -0.4274 & \textbf{0.8131} & 0.3424 \\
     & 16 & \textbf{-0.4080} & 0.8123 & \textbf{0.3421} \\
     & 32 & -0.4748 & 0.7962 & 0.3641 \\
     & 64 & -0.4596 & 0.7978 & 0.3513 \\
     & 128 & -0.4720 & 0.7937 & 0.4647 \\
    \hline
    \end{tabular}
    \end{table}

\section{Conclusion} \label{conclusion}

This paper introduces TPP-TAL, a unified framework designed to enhance temporal awareness in Large Language Models for continuous-time event modeling by integrating two complementary components: the Temporal Cross-Fusion (TCF) and the Multi-Scale Temporal Bias Transformer (MTBT). The proposed model bridges fine-grained temporal–semantic fusion within events and multi-scale temporal reasoning across events. This design enables the LLM to capture both local semantic dynamics and global temporal dependencies within a coherent and interpretable architecture. Experimental results across four diverse real-world datasets demonstrate that TPP-TAL consistently outperforms the strong baseline TPP-LLM in terms of log-likelihood, accuracy, and RMSE, confirming its superiority in modeling event intensity, temporal prediction, and semantic classification. The TCF module enables adaptive temporal conditioning at the token level, effectively aligning asynchronous event semantics, while the MTBT module introduces a hierarchical temporal bias structure that captures long-range and periodic dependencies. Their synergistic interaction equips the LLM with fine-grained temporal sensitivity and robust generalization across heterogeneous temporal patterns.
    
Future work will explore two directions. On the one hand, we will extend TPP-TAL to multimodal temporal modeling, incorporating spatial, visual, or contextual modalities to enhance real-world applicability. On the other hand, we aim to scale the framework to larger LLM backbones and online streaming settings, enabling real-time event reasoning and long-horizon temporal forecasting.

\section*{Data Availability}
Code and datasets are available at 
https://github.com/\\chenlilil/TPP-TAL.

\section*{CRediT Authorship Contribution Statement}
\textbf{Lili Chen}: Methodology, writing original draft.
\textbf{Wensheng Gan}: Review and editing, supervision.
\textbf{Shuang Liang}: Review and editing, supervision.
\textbf{Philip S. Yu}: Review and editing.

\section*{Declaration of Competing Interest}

The authors declare that they have no known competing financial interests or personal relationships that could have appeared to influence the work reported in this paper.


\bibliographystyle{unsrt}

\bibliography{main.bib}

\end{document}